\documentclass[letterpaper, 10 pt, conference]{ieeeconf}

\usepackage{amsmath,amsfonts}
\usepackage{array}
\usepackage{enumerate}
\usepackage[utf8]{inputenc}
\usepackage[T1]{fontenc}
\usepackage{amsfonts}
\usepackage{amssymb}
\usepackage{tabularx}
\makeatletter
\let\NAT@parse\undefined
\makeatother
\usepackage{hyperref}
\hypersetup{
    colorlinks=true,
    linkcolor=blue,
    filecolor=magenta,      
    urlcolor=cyan,
    }
\IEEEoverridecommandlockouts 
\usepackage{algorithm}  
\usepackage{algpseudocode}  
\usepackage{breqn}
\usepackage{caption}
\usepackage{subcaption}
\usepackage[dvipsnames]{xcolor}

\usepackage{textcomp}
\usepackage{stfloats}
\usepackage{url}
\usepackage{verbatim}
\usepackage{graphicx}
\usepackage{cite}
\usepackage{color}
\usepackage{colortbl}
\let\labelindent\relax
\usepackage{enumitem}
\usepackage{tabularray}
\usepackage{multirow}
\usepackage{diagbox}
\usepackage{booktabs}
\usepackage{hhline}

\usepackage{framed}
\usepackage[most]{tcolorbox}
\usepackage{xcolor}
\colorlet{shadecolor}{orange!15}
\usepackage[font=footnotesize]{caption}

\definecolor{graytext}{rgb}{0.483, 0.483, 0.4585}       
\definecolor{purpletext}{rgb}{0.626, 0.439, 0.678}      
\definecolor{greentext}{rgb}{0.417, 0.650, 0.294}        
\definecolor{orangetext}{rgb}{0.673, 0.508, 0.228}       
\definecolor{bluetext}{rgb}{0.327, 0.549, 0.7}           

\definecolor{blacktext}{rgb}{0, 0, 0}

\title{\LARGE \bf REBEL: Rule-based and Experience-enhanced Learning with LLMs for Initial Task Allocation in Multi-Human Multi-Robot Teaming}

\author{Arjun Gupte$^{1}\dag$, Ruiqi Wang$^{1}\dag$, Vishnunandan L.N. Venkatesh$^{1}$, Taehyeon Kim$^{1}$, \\ Dezhong Zhao$^{1,2}$, Ziqin Yuan$^{1}$, and Byung-Cheol Min$^{1}$
\thanks{ $\dag$ Equal Contribution}
\thanks{$^{1}$SMART Laboratory, Department of Computer and Information Technology, Purdue University, West Lafayette, IN, USA. {\tt\small{[guptea, wang5357, lvenkate, kim4435, yuan460, minb]@purdue.edu}.}}
\thanks{$^{2}$College of Mechanical and Electrical Engineering, Beijing University of Chemical Technology, Beijing, China. \tt\small{DZ\_Zhao@buct.edu.cn}.}
\thanks{This paper is based on research supported by the National Science Foundation (NSF) under Grant No. IIS-1846221.}}

\begin{document}
\setlength{\abovedisplayskip}{1pt} 
\setlength{\belowdisplayskip}{1pt} 

\maketitle

\begin{abstract}
Multi-human multi-robot teams are increasingly recognized for their efficiency in executing large-scale, complex tasks by integrating heterogeneous yet potentially synergistic humans and robots. However, this inherent heterogeneity presents significant challenges in teaming, necessitating efficient initial task allocation (ITA) strategies that optimally form complementary human-robot pairs or collaborative chains and establish well-matched task distributions. While current learning-based methods demonstrate promising performance, they often incur high computational costs and lack the flexibility to incorporate user preferences in multi-objective optimization (MOO) or adapt to last-minute changes in dynamic real-world environments. To address these limitations, we propose REBEL, an LLM-based ITA framework that integrates rule-based and experience-enhanced learning to enhance LLM reasoning capabilities and improve in-context adaptability to MOO and situational changes. Extensive experiments validate the effectiveness of REBEL in both single-objective and multi-objective scenarios, demonstrating superior alignment with user preferences and enhanced situational awareness to handle unexpected team composition changes. Additionally, we show that REBEL can complement pre-trained ITA policies, further boosting situational adaptability and overall team performance. Website at \url{https://sites.google.com/view/ita-rebel}. 
\end{abstract}

\section{Introduction}
Multi-human multi-robot (MH-MR) teams are dynamic, heterogeneous groups where multiple humans and robots, each with unique capabilities, collaborate simultaneously to accomplish tasks with varied requirements \cite{dahiya2023survey}. Compared to dyadic human-robot teams, MH-MR teams harness the distinct yet complementary strengths of diverse agents. This synergy has demonstrated significant potential in high-stakes, large-scale domains such as search-and-rescue operations, environmental surveillance, warehouse logistics, and manufacturing assembly lines \cite{teaming2022state, dahiya2023survey}.

While the heterogeneity of MH-MR teams offers greater adaptability and scalability, it also introduces significant coordination challenges \cite{yuan2025adaptive}. An essential first step in addressing these challenges is to establish a teaming strategy that harnesses this heterogeneity to form complementary human-robot pairs or collaborative chains within a specific task context \cite{wang2023initial}. This requires efficient initial task allocation (ITA) strategies that adaptively assign roles, define human-robot collaboration patterns, and initialize task distributions by considering team heterogeneity under different task requirements during the teaming stage. Poor ITA decisions can lead to inefficient team formations, reducing overall performance and limiting the effectiveness of in-operation task reallocation strategies. This, in turn, makes it difficult to restore optimal performance even after prolonged operational adjustments \cite{jo2023affective, wu2022task, wang2024husformer, yuan2025adaptive, patel2020improving, schmidbauer2023empirical}.

\begin{figure}[t]
\centering
\includegraphics[width=1\columnwidth]{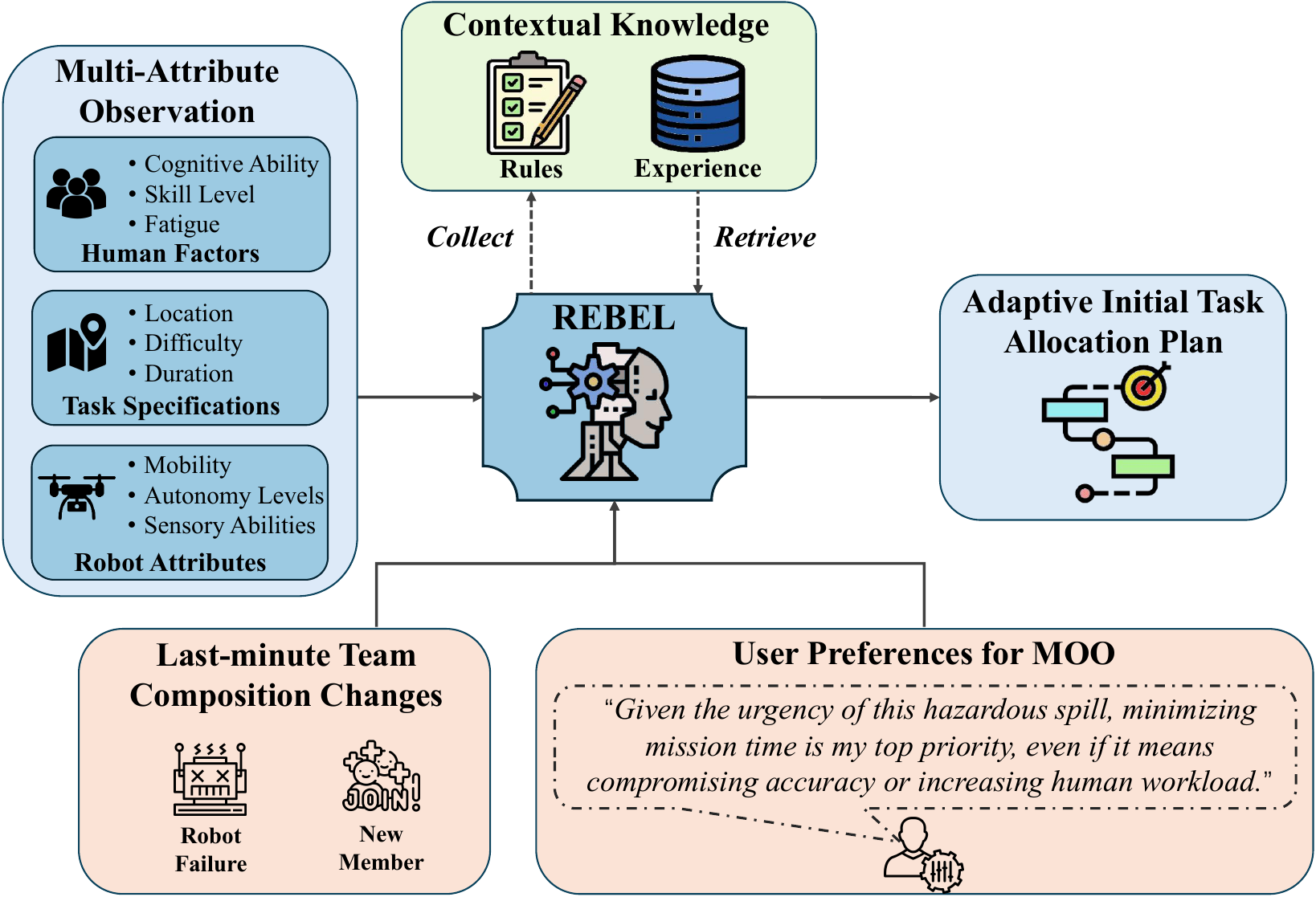}
\vspace{-15pt}
\caption{Conceptual illustration of the proposed LLM-based REBEL framework for ITA in MH-MR teams. Given a multi-attribute observation that reflects the heterogeneity of the MH-MR team, assigned tasks, potential user preferences for multi-objective optimization (MOO), and last-minute team composition changes, the LLM generates an adaptive ITA plan. The system also retrieves the most relevant guidance rules and prior experiences from previous interactions to enhance decision-making.}
\vspace{-15pt}
\label{fig:concept}
\end{figure}

Existing literature on ITA in MH-MR teaming can be broadly categorized into model-based and learning-based approaches. Model-based methods utilize mathematical models, such as mixed-integer linear programming (MILP) \cite{fu2022robust, lippi2023human, chatzikonstantinou2020integrated, zhang2016co, humann2018modeling}, to define decision rules. While these methods offer strong mathematical rigor, they often rely heavily on extensive model engineering, limiting their generalizability and hindering the development of universal rule-based models.

Recently, learning-based approaches \cite{wang2023initial, wang2024initial} have demonstrated greater adaptability by learning through trial and error with diverse task requirements and team configurations. For example, prior work \cite{wang2023initial} formulates ITA as a multi-attribute contextual Markov Decision Process (MDP) and introduces a reinforcement learning (RL) method. Furthermore, \cite{wang2024initial} proposes a hierarchical RL approach to enhance scalability for larger teams and handle more complex tasks. While these methods outperform model-based ones, they still face several key limitations.

Most existing methods focus solely on single-objective optimization (SOO), prioritizing team performance. However, in real-world MH-MR teaming, multiple objectives often extend beyond performance metrics. Factors such as mission time, human workload, and environmental constraints (e.g., operational noise levels) can significantly influence decision-making \cite{humann2023modeling}. Additionally, end-user preferences may vary depending on specific mission requirements \cite{yang2024driving,wang2024prefclm}, necessitating an adaptive multi-objective optimization (MOO) approach.

Furthermore, current methods often lack the flexibility and situational awareness needed to manage last-minute changes in team composition due to unforeseen events such as robot malfunctions, battery depletion, or human unavailability. This limitation stems from the static state space requirements in RL training or the rigid parameter dependencies in model-based methods, both of which restrict adaptability across different scenarios. Consequently, the need to retrain models for each new team configuration or scenario not only incurs significant computational costs but also hinders practical deployment in dynamic, real-world environments.

To address these challenges, as shown in Fig.~\ref{fig:concept}, we propose REBEL, an LLM-based framework for ITA in MH-MR teaming. To enhance the reasoning capabilities of LLMs and in-context flexibility, REBEL employs a rule-based, experience-enhanced approach. Specifically, during the knowledge acquisition phase, LLMs generate, collect, and refine ITA rules and experiences for each objective by testing various ITA strategies under different simulated scenarios and accumulating performance data. This performance data includes task allocation outcomes, such as success rates, execution times, and other relevant mission-specific indicators.

For inference in new scenarios, REBEL utilizes a retrieval-augmented generation (RAG) mechanism to dynamically retrieve instructional rules and prior experiential data for both SOO and MOO. This retrieved context serves as additional input for reasoning, enabling in-context learning (ICL) without the need for costly fine-tuning. By combining rule learning with experience retrieval, REBEL effectively handles complex user preferences in MOO and adapts allocations to last-minute changes in team composition, leveraging accumulated knowledge from prior deployments to maintain optimal performance. The primary contributions of this work are summarized as follows: 
\begin{itemize}[leftmargin=*]
    \item We propose an LLM-based framework for ITA in MH-MR teams as a deployment-efficient few-shot approach. It enhances stakeholder preference adaptation in MOO and is more flexible to last-minute changes in team composition.
    \item We introduce a rule-based, experience-enhanced learning module that improves the in-context adaptability of LLMs in MOO and dynamic situational changes by leveraging accumulated knowledge from prior deployments without costly re-training or fine-tuning.
    \item Our method not only functions independently but also complements pre-trained RL-based ITA policies, enhancing their situational awareness and overall performance.
    \item  We conduct extensive evaluations in benchmark simulation environments to validate our approach across various settings.
\end{itemize}

\section{Background and Related Works}

\subsection{ITA in MH-MR Teaming}
ITA in MH-MR teaming aims to adaptively define roles, establish collaboration patterns between humans and robots (e.g., fully autonomous vs. shared control), and initialize the distribution of tasks to best match individual capabilities across team members \cite{wang2023initial}. The fundamental objective of ITA is to maximize team performance \cite{wang2024initial}. However, in real-world scenarios, ITA often extends beyond a single objective, evolving into a MOO problem, where stakeholders may prioritize different objectives based on specific mission requirements, such as mission time, human workload, and safety considerations \cite{humann2023modeling}.


Adding to the complexity is the fact that team composition is often not static, with changes potentially occurring even at the last minute \cite{dahiya2023survey}. For example, unexpected robot failures, human unavailability due to emergencies, or sudden shifts in mission requirements can all disrupt preset teaming plans. Therefore, an effective ITA strategy requires situational awareness to adapt to such changes dynamically while still making optimal task allocation decisions.

However, existing works may not efficiently manage the aforementioned real-world ITA dynamics. Most methods \cite{fu2022robust, mina2020adaptive, lippi2023human, chatzikonstantinou2020integrated, zhang2016co, wang2023initial,wang2024initial} focus exclusively on SOO, prioritizing team performance while neglecting user preferences. While it is possible to add more optimization factors to model-based approaches or apply multi-objective reward shaping in learning-based methods, these solutions are often rigid and lack flexibility. More importantly, both model-based \cite{fu2022robust, mina2020adaptive, lippi2023human, chatzikonstantinou2020integrated, zhang2016co, humann2023modeling} and learning-based approaches \cite{wang2023initial,wang2024initial} struggle to adapt to new scenarios, particularly when faced with unexpected last-minute changes in team composition or mission requirements due to their reliance on fixed state inputs. This inability to generalize dynamically hinders situational adaptability, leading to suboptimal ITA decisions.

In contrast, our method leverages the flexibility and generalization capabilities of LLMs, integrating an in-context learning module to better address user preferences in MOO and enhance situational awareness to last-minute team changes in dynamic and unpredictable environments.

\subsection{LLMs for Robotics and Task Allocation}
Recently, LLMs have shown promising potential in robotics domains such as task allocation in multi-robot systems \cite{kannan2024smart, talebirad2023multi} and user intention perception in human-robot interaction \cite{wang2024prefclm, liang2023code}. These advancements suggest that LLMs could offer a viable solution to address the aforementioned ITA challenges in MH-MR teaming. However, existing LLM-based approaches for task allocation and coordination are predominantly limited to SOO \cite{kannan2024smart, talebirad2023multi,liang2023code}, focusing primarily on task performance. Such approaches may struggle to adapt to more complex MOO scenarios, where the uncertainty and inconsistency of LLM reasoning can significantly increase, potentially leading to suboptimal ITA decisions in dynamic environments.

In contrast to these LLM-based allocation methods, we propose a novel ICL module within REBEL, which combines rule-based reasoning with experience-enhanced learning. This approach not only supports MOO by reflecting user preferences, but also enhances situational awareness in dynamic ITA scenarios. By collecting and refining single-objective-based rules and experiences and leveraging RAG for MOO inferencing, our method dynamically incorporates relevant rules and prior experiences, enabling the LLM to maintain consistency and adaptability even in unpredictable mission contexts.


\begin{figure*}[t]
\centering
\includegraphics[width=0.9\linewidth]{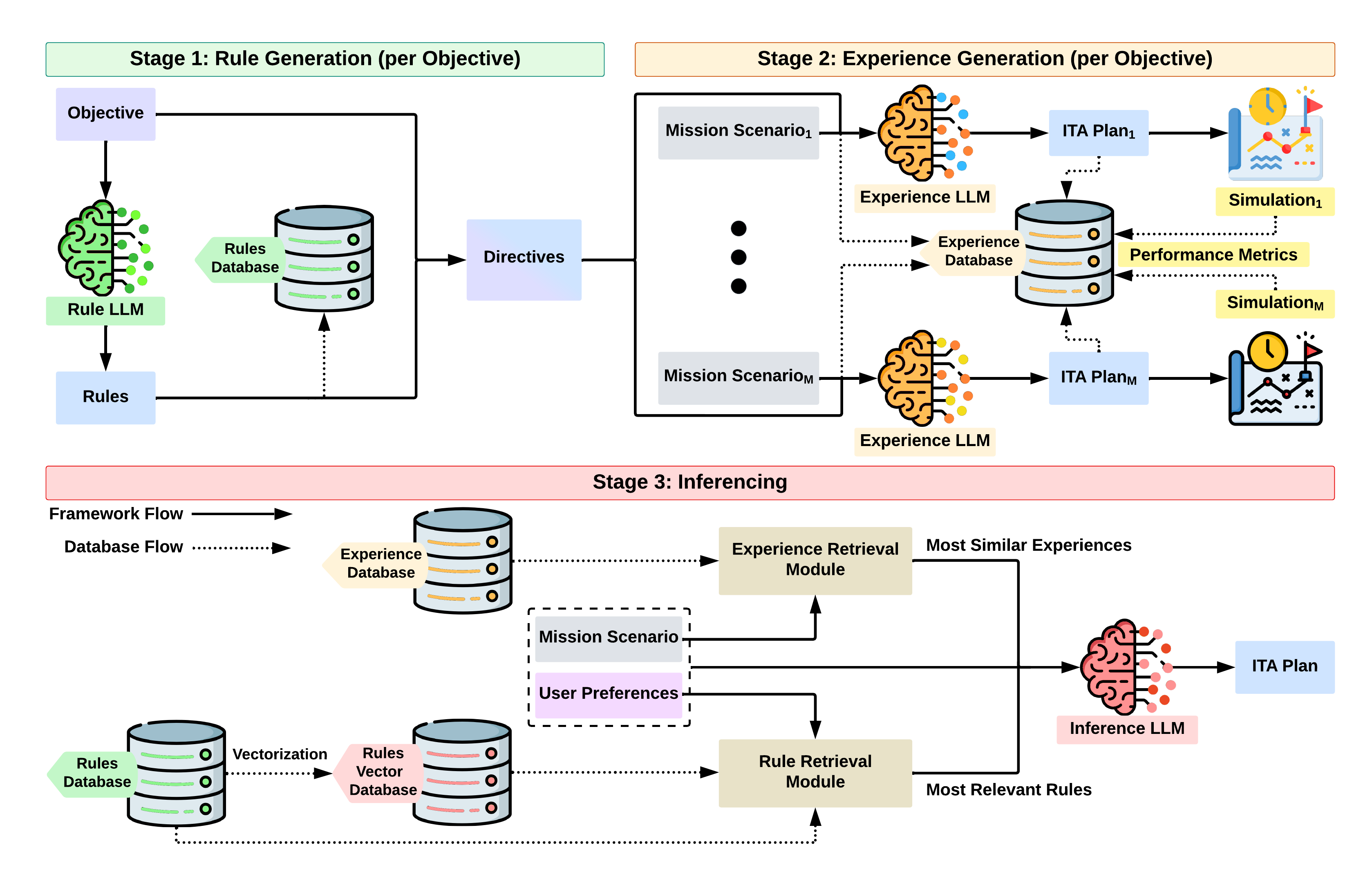}
\vspace{-16pt}
\caption{Illustration of the three stages in the proposed LLM-based REBEL framework for ITA in MH-MR teams. The first two stages comprise the Knowledge Acquisition phase in which the LLM creates ITA plans for different randomized missions and generates a collection of learned rules and experience data through simulation. During the Inferencing stage, the LLM leverages the Rule and Experience Retrieval modules to extract rules and experiences most relevant to the user's input to enhance the quality of its ITA plan.}
\vspace{-15pt}
\label{fig:framework}
\end{figure*}

\section{Methodology}
\subsection{Problem Formulation}
Consider an MH-MR team comprising multiple human operators $\mathcal{H} = \{H_1, H_2, \ldots, H_n\}$ and a fleet of robots $\mathcal{R} = \{R_1, R_2, \ldots, R_m\}$. Each agent \(a \in \mathcal{H} \cup \mathcal{R}\) is characterized by a capability profile \(\mathrm{Cap}(a)\). This profile is measured across multiple dimensions, such as cognitive ability and operational skill level for humans and mobility and sensor quality for robots. The team's mission consists of a set of tasks $\mathcal{T} = \{T_1, T_2, \ldots, T_k\}$, where each task \(T\) has specific requirements represented by a requirement vector \(\mathrm{Req}(T_i)\). This vector quantifies task specifications across multiple aspects, such as spatial location and complexity.

The ITA decision is defined as a mapping from each task to a high-level action list, specifying the assigned agents and their collaboration patterns. Formally, for a given task \(T_i\), the allocation function \(f\) returns a set of action decisions: $f(T_i) = \{(a_1, p_1), (a_2, p_2), \dots, (a_q, p_q)\}$
where each \(a_j\) is an agent (human or robot) and each \(p_j\) denotes the collaboration pattern or execution mode (e.g., \texttt{human-only}, \texttt{robot-autonomous}, or \texttt{shared control}). The joint ITA decision $\mathcal{A}$ for the entire mission is the aggregation of high-level action lists across all tasks: \(\mathcal{A} = f(T_1) \times f(T_2) \times \cdots \times f(T_k)\).

The optimization goal is multi-objective, defined as: \(\mathcal{O} = \{O_1,\, O_2,\, \ldots,\, O_p\}\).  In our case study, we consider three objectives: task performance, mission completion time, and human workload, although the framework is generalizable to incorporate additional objectives as needed. User preferences are integrated through a weight vector: \(\boldsymbol{\lambda} = (\lambda_1,\, \lambda_2,\, \ldots,\, \lambda_p)\) where each element $\lambda_i \in [0, 1]$, resulting in an aggregate objective function:

\vspace{-7pt}
\begin{equation}
   J(\mathcal{A}) = \sum_{j=1}^{p} \lambda_j\, O_j\bigl(\mathcal{A}, \mathcal{H}, \mathcal{R}\bigr) 
\end{equation}

The ITA problem is then formulated as the following constrained optimization:
\begin{equation}
\resizebox{.91\columnwidth}{!}{$
\begin{aligned}
\max_{\mathcal{A}} \quad & J(\mathcal{A}) \\
\text{s.t.} \quad & \forall T_i \in \mathcal{T}, \quad \bigcup_{(a_j, p_j) \in f(T_i)} \mathrm{Cap}(a_j, p_j) \supseteq \mathrm{Req}(T_i)
\end{aligned}$}
\end{equation}

This formulation ensures that task allocations satisfy all requirements while optimizing multi-objective trade-offs based on user-defined preferences. To solve this problem, we introduce REBEL, which leverages rule-based reasoning and experience-enhanced learning to improve LLM-driven ITA adaptability. The detailed methodology and implementation of REBEL are presented in the following sections.

\subsection{Framework Overview}
As illustrated in Fig.~\ref{fig:framework}, REBEL operates through three sequential stages: Rule Generation, Experience Generation, and Inferencing. The first two stages constitute the Knowledge Acquisition (KA) phase, where objective-specific rules are constructed and experience data is collected through few-shot mission simulations, recording relevant performance metrics. These stages are executed jointly for each optimization objective in $\mathcal{O}$ to ensure comprehensive knowledge acquisition. The KA phase can be conducted prior to deployment through simulated missions and continuously refined during long-term development by incorporating real-world experiences from past deployments.

During the Inferencing stage, REBEL is presented with an unseen mission scenario and tasked with performing ITA tailored to user-defined preferences. To improve decision-making, REBEL uses retrieval-augmented generation (RAG) to retrieve relevant rules and past experiences \cite{lewis2020retrieval}. This enables the system to adapt dynamically and generalize effectively across varying operational conditions. 

To ensure that the LLM effectively comprehends contextual information, we introduce a Structured Prompting Format (SPF), inspired by \cite{singh2023progprompt, kannan2024smart}, to format all the information and prompts the LLMs require throughout all stages.



\subsection{Stage 1: Rule Generation}
The Rule Generation stage employs an LLM to generate prescriptive, objective-specific rules that guide the ITA process during Stages 2 and 3. Given a set of objectives $\mathcal{O}$, the LLM is iteratively prompted to produce a structured set of rules tailored to each objective. To facilitate this, the prompt for the LLM is formulated as: $\mathcal{P} = (B, G, O_j)$. $\mathcal{P}$ contains the first three components of the SPF, where $B$ represents background information and informs the LLM of the format in which the team composition will be provided. $G$ is the task to be performed and $O_j$ is the current objective. A truncated example of the SPF-compliant prompt is shown below:

\begin{lstlisting}
[black]Background
  Human Attributes: {H_#: [Skill, Cognition], ...}
  Robot Details: {R_#: [Speed, Camera Quality], ...}
  Task Info: {T_#: [(x, y), Difficulty], ...}[/black]
[purple]Mission Objectives
  Minimize the overall mission time.[/purple]
[blue]Goal
  Generate a set of rules to follow during ITA.[/blue]

\end{lstlisting}

As shown in Fig.~\ref{fig:framework}, the set of generated rules across all $p$ objectives, $\Gamma = \{\gamma_1, \ldots, \gamma_p\}$, is stored as plain text in a Rules Database $\mathcal{D}_\Gamma$. The LLM can also iteratively refine the rules for each objective by observing the experience data introduced below. 




\subsection{Stage 2: Experience Generation} 
The goal of the Experience Generation stage is to enable the LLM to gain practical experience in applying the rules generated in Stage 1 in different scenarios. The LLM acquires practical knowledge in a few-shot manner, by performing ITA for several missions per objective. Each mission is configured randomly, with different team sizes, team member characteristics, and task attributes specified in the LLM's prompt. The rules generated in Stage 1 are provided to the LLM when missions with the corresponding objective are executed in simulation.

At this point, the prompt is represented as: $\mathcal{P} = (S, G, O_j, \gamma_j)$. The first three components mirror those of Stage 1. However, $\gamma_j$, the rules generated specifically for the current objective $O_j$ are retrieved from $\mathcal{D}_\Gamma$ and added to the prompt, completing the fourth component of the SPF. This combination of rules and the objective is represented as the Directives in Fig.~\ref{fig:framework}. A truncated sample of the prompt is shown below:

\begin{lstlisting}
[gray]Mission Scenario
  Human Attributes: {H_0: [Lo, Med], H_1: [Hi, Lo]}
  Robot Details: {UAV_0: [13, Lo], UGV_0: [6, Med]}
  Task Info: {T_0: [(9, 5), Hi], T_1: [(2, 7), Lo]}[/gray]
[blue]Goal
  Perform ITA for the provided mission.[/blue]
[purple]Mission Objectives
  Minimize the overall mission time.[/purple]
[green]Rules
    Assign faster robots to tasks farther away.
    ...[/green]
\end{lstlisting}

An example of an LLM-generated ITA plan for this particular mission scenario is as follows:

\begin{lstlisting}
[black]T_0: (H_1, UAV_0)
T_1: (H_0, UGV_0)[/black]
\end{lstlisting}

REBEL parses this plan and deploys the human and robot agents in the mission accordingly in a simulation environment. For each of the $k$ missions simulated for $O_j$, performance data \( \pi^k_j = (A_m, T_m, U_h) \) is collected, where $A_m$ is the mission accuracy, $T_m$ is the mission duration, and $U_h$ is the average human utilization. Similar to Stage 1, the performance data across all $p$ objectives is defined as: $\Pi = \{\pi_1, \ldots, \pi_p\}$. Mission Scenarios, LLM-generated ITA plans, and corresponding performance data are stored as plain text in an Experience Database $\mathcal{D_E}$ as shown in Fig.~\ref{fig:framework}. By learning from these experiences, the LLM can adjust its strategies to improve ITA efficiency in dynamic and unpredictable environments during Inferencing.




\subsection{Stage 3: Inferencing} 
During Inferencing, REBEL is tasked with performing ITA on an unseen mission while leveraging the rules and experience data previously collected during KA. The user gives REBEL an instruction via the prompt $\mathcal{P} = (S, G, \Theta)$. MH-MR teams may be deployed in situations that require a careful balance of multiple potentially conflicting user preferences. To reflect the multifaceted nature of real-world missions, $\Theta$ can be a single objective $O_j$ or a set of objective-weight pairs $\{(O_1, \lambda_1), (O_2, \lambda_2), \ldots, (O_n, \lambda_n)\} \in \mathcal{O}$ to reflect multiple user preferences.

We introduce two retrieval modules as seen in Fig.~\ref{fig:framework}. The Rule Retrieval module retrieves the most relevant rules for $\Theta$ using an Ensemble Retriever \cite{langchainEnsembleRetriever}. The Ensemble Retriever is a hybrid of a Sparse Retriever that utilizes the Best Match 25 (BM25) keyword matching algorithm \cite{robertson2009probabilistic} and a Dense Retriever that utilizes more complex semantic similarity techniques when searching the Rules Vector Database. By combining the strengths of these two retrieval methods, the Ensemble Retriever ensures that the most contextually appropriate rules are retrieved. The BM25 Sparse Retriever focuses on keyword similarity, as shown in Eqs.~\ref{eq:bm25} and~\ref{eq:idf}:
\vspace{3pt}
\begin{equation}
\resizebox{.9\columnwidth}{!}{$
\text{Sparse}(\Theta, \gamma^k_j) = \sum_{t \in \Theta} \text{IDF}(t) \cdot \frac{\text{TF}(t, \gamma^k_j) \cdot (k_1 + 1)}{\text{TF}(t, \gamma^k_j) + k_1 \cdot (1 - b + b \cdot \frac{|\gamma^k_j|}{avg(\gamma)})}$}
\label{eq:bm25}
\end{equation}
\vspace{-4pt}

\begin{equation}
\text{IDF}(t) = \log\left(\frac{N - n(t) + 0.5}{n(t) + 0.5}\right)
\label{eq:idf}
\end{equation}

Here, Sparse$(\Theta, \gamma^k_j)$ represents the keyword similarity score between the user preferences and an individual rule $\gamma^k_j$ within $ \mathcal{D}_\Gamma \; \forall j \in \{1, 2, \ldots, p\}, \forall k \in \{1, 2, \ldots, n_j\}$ where $n_j$ is the number of rules generated for objective $O_j$. $t$ is a token, IDF$(t)$ is the Inverse Document Frequency, and TF$(t, \gamma^k_j)$ is the Term Frequency of $t$ in $\gamma^k_j$.
Furthermore, $N$ is the total number of rules in $\mathcal{D}_\Gamma$, $n(t)$ is the number of rules containing the token $t$. 
Finally, $|\gamma^k_j|$ is the number of tokens in $\gamma^k_j$, $avg(\gamma)$ is the average number of tokens across all rules in $\mathcal{D}_\Gamma$, and $k_1$ and $b$ are the term frequency saturation and length normalization hyperparameters. 

The Dense Retriever identifies deeper-level semantic similarity using cosine similarity and computes a similarity score $\text{Dense}(\Theta, \gamma^k_j)$ between the text embeddings as follows:

\vspace{-6pt}
\begin{equation}
\text{Dense}(\Theta, \gamma^k_j) = \frac{\boldsymbol{\Theta} \cdot \boldsymbol{\gamma^k_j}}{||\boldsymbol{\Theta}|| \cdot ||\boldsymbol{\gamma^k_j}||}
\label{eq:dense_rtrvr}
\end{equation}
\vspace{-8pt}

Each retriever ranks all scores and stores these rankings in lists $\mathcal{L}_{sparse}$ and $\mathcal{L}_{dense}$. The two lists are fused in a proportion $\alpha = 0.5$ with hyperparameter $c$ using Reciprocal Rank Fusion \cite{cormack2009reciprocal}, yielding the most relevant rules $\gamma^*$:

\vspace{-5pt} 
\begin{equation}
\resizebox{.9\columnwidth}{!}{$
\gamma^* = \text{TopRanked}_{\gamma^k_j \in \mathcal{D}_\Gamma} \left[\frac{\alpha}{c + \mathcal{L}_{sparse}(\gamma^k_j)} + \frac{1 - \alpha}{c + \mathcal{L}_{dense}(\gamma^k_j)}\right] $}
\label{eq:ensemble_rtrvr}
\end{equation}
\vspace{-2pt} 

The Experience Retrieval module retrieves the most similar missions from $\mathcal{D_E}$, given the current Mission Scenario $S$. To enhance the retrieval process, we develop a custom algorithm that separates $S$ into multiple components and performs a similarity search for each as outlined below: 

\begin{enumerate}
    \item $S$ is divided into three dictionaries: Human Attributes, Robot Details, and Task Info. These sections are represented as $H_{dict}$, $R_{dict}$, and $T_{dict}$, respectively.

    \item Each dictionary is converted to a text embedding using a BERT-like model \cite{devlin2019bert}, resulting in $\mathbf{h}$, $\mathbf{r}$, and $\mathbf{t}$.

    \item Similarly, the Mission Scenarios stored in $\mathcal{D_E}$ are divided into three sections and embedded accordingly, where $\mathbf{H} = \{\mathbf{h}_1, \mathbf{h}_2, \ldots, \mathbf{h}_e\}, \mathbf{R} = \{\mathbf{r}_1, \mathbf{r}_2, \ldots, \mathbf{r}_e\}, \mathbf{T} = \{\mathbf{t}_1, \mathbf{t}_2, \ldots, \mathbf{t}_e\}$ and $e$ is the total number of Mission Scenarios in $\mathcal{D_E}$.
    
    \item A cosine similarity score $C$ is calculated between the embedding of $S$ and each mission in $\mathcal{D_E}$ to retrieve the $k$ most similar instances $\mathbf{H^*}$, $\mathbf{R^*}$, $\mathbf{T^*}$:

\end{enumerate}
\begin{equation}
\resizebox{.91\columnwidth}{!}{$
\mathcal{I}^* = \operatorname{Top}_k\left\{ i \in \{1, 2, \ldots, e\} : C_{\mathbf{h},\mathbf{H}[i]} + C_{\mathbf{r},\mathbf{R}[i]} + C_{\mathbf{t},\mathbf{T}[i]} \right\}$}
\end{equation}
\vspace{-10pt}
\begin{equation}
\resizebox{.91\columnwidth}{!}{$
\mathbf{H}^* = \{ \mathbf{H}[i] : i \in \mathcal{I}^* \}, \quad \mathbf{R}^* = \{ \mathbf{R}[i] : i \in \mathcal{I}^* \}, \quad \mathbf{T}^* = \{ \mathbf{T}[i] : i \in \mathcal{I}^* \}$}
\end{equation}
\vspace{-17pt}

Finally, these selected missions are re-ranked based on their performance in achieving $\Theta$. The top missions and their corresponding ITA plans are added as few-shot examples to the LLM prompt. The prompt now consists of all components of the SPF, as can be seen in the following:

\begin{lstlisting}
[gray]Mission Scenario
  Human Attributes: {H_0: [Lo, Med], H_1: [Hi, Lo]}
  Robot Details: {UAV_0: [13, Lo], UGV_0: [6, Med]}
  Task Info: {T_0: [(9, 5), Hi], T_1: [(2, 7), Lo]}[/gray]
[blue]Goal
  Perform ITA for the provided mission.[/blue]
[purple]Mission Objectives
  Minimize mission time (weight = 0.55)
  Maximize mission accuracy (weight = 0.35)
  Minimize human workload (weight = 0.10)[/purple]
[green]Rules
  Assign faster robots to tasks farther away.
  Assign skilled humans to difficult tasks.
  ...[/green]
[orange]Prior Experience
  Mission Scenario + ITA Plan A
  Mission Scenario + ITA Plan B
  ...[/orange]
\end{lstlisting}

By integrating both rule-based learning and experiential data, the LLM can adapt to new scenarios and effectively perform SOO and MOO with improved situational awareness, resulting in higher ITA performance.

\section{Case Study and Experiments}
\label{case}

\subsection{Task Scenario and Configuration}
\label{caseset}
We designed a case study involving an extensive environmental surveillance task centered on monitoring pollution leaking from a warehouse. Pollutants originating from the warehouse require precise observation and identification by the MH-MR team. The mission begins when a satellite system detects and catalogs various points of interest (POIs). The team is tasked with two primary objectives: 1) Moving to each POI to capture images and 2) Analyzing the acquired images to determine whether a POI represents an actual hazard. We use the same human factors, robot characteristics, and task specifications from \cite{wang2024initial}, and they are summarized as: i) Human Factors\cite{harriott2013modeling}: Cognitive ability, Operational skill level; ii) Robot Characteristics: Mobility (speed), Sensory capabilities (camera quality) in UGVs and UAVs, Operation mode (fully autonomous or under collaborative control with human operators); and iii) Task Specifications: Locations of the POIs, Intrinsic complexity levels.

\subsection{Simulation Environment}
\begin{figure}[t]
\centering
\includegraphics[width=0.98\columnwidth]{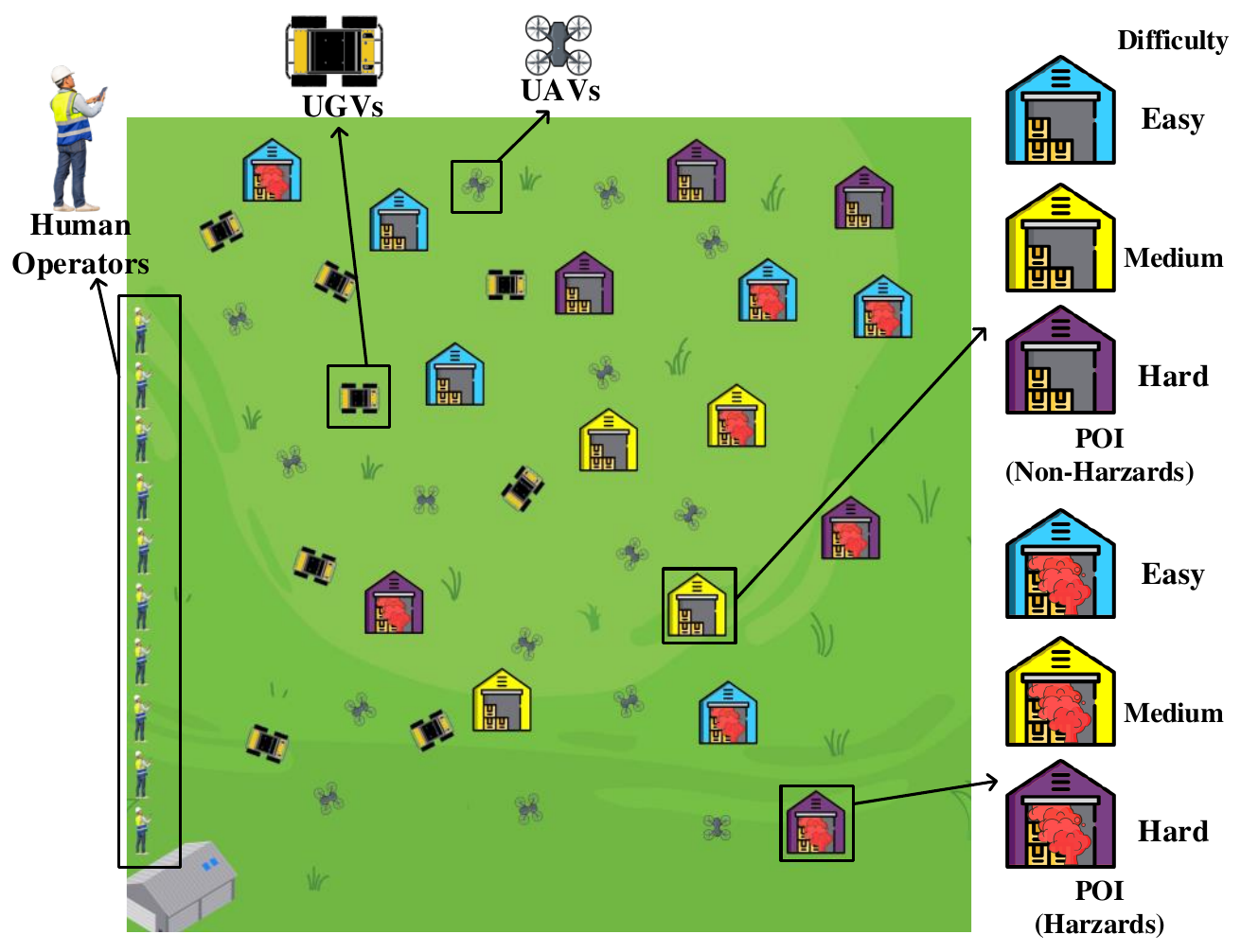}
\vspace{-5pt}
\caption{A visual representation of the simulation environment is depicted. The scale has been adjusted to enhance clarity and visualization. Each POI is distinguished by color to indicate the complexity level for hazard evaluation.}
\label{fig:sim}
\vspace{-15pt}
\end{figure}

We utilized a benchmark simulation environment for ITA in MH-MR teams \cite{wang2024initial}. The environment spans a $2~km \times 2~km$ area requiring hazard surveillance by an MH-MR team and is populated with multiple POIs. Each POI corresponds to a building, and the color of each building indicates its inherent complexity level for hazard evaluation. A visual representation of this environment is provided in Fig.~\ref{fig:sim}.

\subsubsection{Robot Model}
Both the speed of the robots and the quality of the images they capture depend on the control mode. When robots navigate autonomously to a POI, they use default parameters. In shared control mode with humans, these parameters are adjusted based on the operator's level of expertise. Additionally, the probability of successful onboard hazard classification by the robot is influenced by both the captured image quality and the inherent difficulty of the hazard. The specific attributes of the robots are the same as those in the settings of \cite{wang2024initial}.



\subsubsection{Human Model}
We model human operators as sequential event handlers, where their performance in hazard classification is influenced by fatigue, workload, and task complexity. These factors are moderated by individual cognitive abilities and operational skill levels, as described in \cite{watson2017informing}. The probability of accurate classification of hazards is determined non-linearly, combining these elements according to the model in \cite{humann2018modeling}. Factors such as fatigue and workload are modeled with decreasing performance over time, while task complexity is captured using a sigmoid function to reflect its non-linear impact on performance \cite{pew1969speed}. Cognitive capacity and operational expertise, which affect performance adjustments, are categorized into three tiers (low, medium, high), with values randomly assigned within these ranges for each experiment \cite{humann2018modeling, humann2023modeling}. The simulated human operator behavior described above precisely follows well-established modeling techniques leveraged in human-robot teaming literature. The detailed equations for this human model can be found in \cite{wang2024initial}.

\subsection{Experimental Setting}

\subsubsection{Single Objective Optimization}
We first evaluated REBEL's performance in SOO settings. Three key objectives were independently assessed: Maximizing Task Performance (TP), Minimizing Mission Time (MT), and Minimizing Human Workload (HW), objectives commonly prioritized in real-world human-robot teaming applications \cite{humann2018modeling}.

For comparison, we implemented three baseline methods: vanilla LLM Zero-shot, a model-based approach MILP \cite{humann2023modeling}, and a learning-based state-of-the-art method AtRL \cite{wang2023initial}. For LLM methods, switching between objectives required only prompt modifications, whereas the AtRL baseline necessitated reward shaping. For maximizing TP, we used the original reward function; for minimizing MT, we modified the reward to penalize longer completion times; and for minimizing HW, we adjusted the reward to discourage excessive human task allocation. We followed the original training protocols to ensure optimal RL performance.

We also evaluated two REBEL ablations: one without rule-based learning and another without experience-based enhancement. All experiments simulated an MH-MR team comprising 5 humans, 7 robots, and 30 POIs, with performance averaged across 100 trials per objective.

\subsubsection{Multi-Objective Optimization}
Furthermore, we conducted experiments in MOO settings combining the three objectives from SOO to assess each method's capability to interpret complex user preferences. We adapted the same baselines from the SOO experiments, along with the best-performing REBEL ablation. 

We considered prioritizing TP, MT, and HW separately. In each test, when one objective was prioritized over the others, weights were assigned to emphasize these preferences (e.g., 0.5 for the highest-priority objective and 0.25 for each of the others). We recorded average performance across all three objectives to analyze behavioral patterns under varying user-preference weights.

\subsubsection{Situational Awareness}
We further assessed REBEL's adaptability to dynamic environments by evaluating its situational awareness in response to last-minute changes in team composition. We considered two test scenarios: operations with fewer team members and operations with additional members during execution. For baselines, both model-based and learning-based approaches cannot natively accommodate last-minute variations in team structure due to their fixed input dimensionality constraints. Therefore, we used Zero-shot LLM as our primary baseline, and developed a new hybrid approach called REBEL+AtRL that modifies the task allocation actions generated by pre-trained AtRL models using REBEL's adaptive reasoning mechanism. This integration enabled us to quantify REBEL's ability to enhance the situational awareness and flexibility of conventional RL models when faced with dynamic team configurations not encountered during training.

\section{Results and Analysis}
\subsection{Results of Single Objective Optimization}

Table I summarizes the results from the SOO experiments. REBEL consistently outperforms the vanilla LLM baseline and its ablations (No Exp/No Rule), underscoring the importance of our Experience-enhanced Rule-based Learning approach. Without these components, the LLM lacks critical external knowledge, reducing its effectiveness in making ITA decisions. This highlights the inherent complexity of the ITA task, even when optimizing for a single objective. REBEL also consistently outperforms MILP, which relies on tuned parameters, unlike REBEL's dynamic iterative learning. REBEL performs better than AtRL in minimizing Mission Time, but lags behind in Task Performance and Human Workload. Despite this, the performance gap is relatively small, with AtRL achieving only 9.28\% better Mission Performance and 3.5\% lower Human Workload.

\begin{table}[h]
\newcommand{\highlight}[1]{\textcolor{orange}{\textbf{#1}}}
\centering
\caption{Performance comparison of ITA methods across three single-objective optimization (SOO) scenarios. \textuparrow~ indicates higher values are better, \textdownarrow~indicates lower values are better. Orange highlights indicate best results and those within 10\% of the best.}
\resizebox{\columnwidth}{!}{
\begin{tabular}{l|c|c|c} 
\toprule
\textbf{Method} & \textbf{\begin{tabular}[c]{@{}c@{}}SOO1\\\textuparrow Task Performance\end{tabular}} & \textbf{\begin{tabular}[c]{@{}c@{}}SOO2\\\textdownarrow Mission Time\end{tabular}} & \textbf{\begin{tabular}[c]{@{}c@{}}SOO3\\\textdownarrow Human Workload\end{tabular}} \\ 
\midrule
MILP & 101.60 & 1857.20 & 0.173 \\
AtRL & \highlight{137.90} & 1523.90 & \highlight{0.141} \\ 
\midrule
Zero-Shot & 102.80 & 2088.47 & 0.162 \\
REBEL-No Exp & 121.70 & 1964.42 & 0.168 \\
REBEL-No Rule & 118.32 & 1825.81 & 0.153 \\
REBEL & \highlight{125.10} & \highlight{1495.40} & \highlight{0.146} \\
\bottomrule
\end{tabular}}
\vspace{-12pt}
\label{tab:performance_comparison}
\end{table}

These results demonstrate that REBEL can achieve near state-of-the-art performance within 10\% of advanced RL methods, even without formal training. By leveraging structured integration of relevant rules and experience data RAG, REBEL enables effective in-context learning, leading to more adaptive and optimized ITA plans.

Notably, the training process for AtRL required approximately 24 hours on an NVIDIA A100 GPU, whereas REBEL achieves comparable performance using only 10 examples of prior mission experience data per objective, highlighting its superior sample efficiency. This efficiency, combined with its ability to generalize without extensive retraining, positions REBEL as a strong candidate for complex ITA tasks, particularly in scenarios with limited training data or constrained computational resources.

\begin{figure*}[t]
\centering
\includegraphics[width=\linewidth]{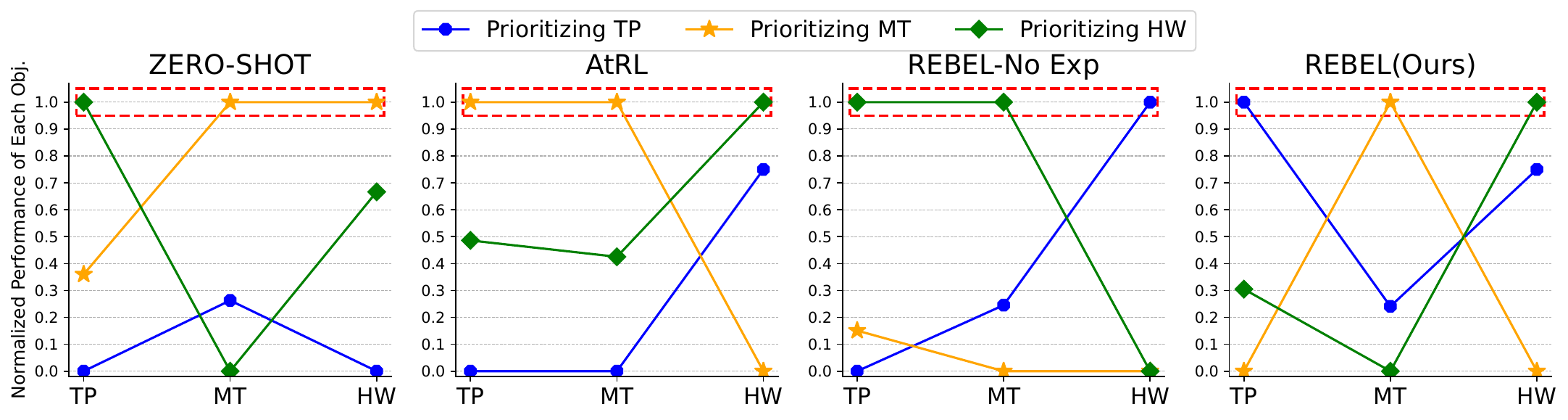}
\vspace{-15pt}
\caption{Performance of different methods in MOO settings in terms of the normalized performance of each objective. TP denotes task performance, MT represents mission time, and HW indicates human workload. The normalized values of each objective under different user preference types are connected by lines to illustrate the preference alignment levels. Blue lines indicate user preference prioritizing TP (with TP, MT, HW weights of 0.5, 0.25, 0.25), orange lines prioritize MT (with weights of 0.25, 0.5, 0.25), and green lines prioritize HW (with weights of 0.25, 0.25, 0.5). The actual prioritized objectives for each method are enclosed in a dotted red box.}
\label{fig:curve}
\vspace{-15pt}
\end{figure*}

\subsection{Results of Multi-Objective Optimization}
Fig.~\ref{fig:curve} showcases the performance of different methods in MOO settings, with normalized values for each objective: task performance (TP), mission time (MT), and human workload (HW), to illustrate preference alignment. The blue, orange, and green lines represent scenarios where TP, MT, and HW were prioritized by users, respectively. The objectives actually prioritized by each method are highlighted in a dotted red box.

From the results, we observe that REBEL exhibits superior sensitivity to user-specified objective preferences compared to the baselines. When a particular objective is prioritized, REBEL consistently aligns the performance of that objective with the user's preferences, as demonstrated by the highest normalized performance value (1.0) in the blue, orange, and green lines for TP, MT, and HW, respectively. This alignment is particularly clear in the red boxed regions in Fig.~\ref{fig:curve}, where REBEL meets the user-defined goals effectively.

In contrast, AtRL only successfully prioritizes MT and HW, but struggles with TP, as indicated by the lower values for this objective. Zero-Shot performs well only for MT, while REBEL-No Exp shows inconsistent preference alignment across all three objectives. This discrepancy highlights the value of incorporating experience data, particularly in more complex multi-objective scenarios, where approaches like REBEL benefit from having additional context and history to interpret and prioritize conflicting objectives.
Moreover, the results reveal that REBEL is more adept at making trade-offs among objectives based on user preferences, compared to the other methods, which exhibit less flexibility in adjusting to shifting priorities. This ability to dynamically reallocate resources and align with user preferences is a critical advantage in real-world MH-MR tasks, where multiple objectives often need to be balanced simultaneously.

\begin{table}[th]
\centering
\caption{Comparison of Task Performance to evaluate situational awareness in two settings.}
\label{tab:sa_performance}
\begin{tabular}{c||cc} 
\toprule
\textbf{Method} & \textbf{Fewer Team Members} & \textbf{More Team Members} \\ 
\midrule
Zero-Shot       & 80.75                       & 87.55
  \\
REBEL           & 106.01                      & 107.90                    \\ 
REBEL + AtRL    & 129.68                      & 158.89                    \\ 
\bottomrule
\end{tabular}
\vspace{-7pt}
\end{table}

\subsection{Results of Situational Awareness}

As shown in Table II, REBEL demonstrates greater robustness under team composition changes, whether increasing or decreasing the number of members, compared to the Zero-Shot LLM approach. Although performance drops slightly from the static setting, REBEL still delivers strong results, highlighting the effectiveness of the proposed rule-based learning and experience-enhanced approach. This adaptability is not only beneficial for SOO and MOO, but also proves advantageous when handling last-minute changes to team composition.

More importantly, REBEL+AtRL shows promising outcomes, illustrating REBEL's potential as a test-time adaptation module for pre-trained RL-based ITA policies. This combination of REBEL's adaptability with AtRL's Reinforcement Learning strengths enhances performance and flexibility, making it a powerful approach for handling real-time team composition changes in dynamic environments.

\section{Conclusion and Future Work}
In this work, we introduced REBEL, a novel LLM-based framework for ITA in MH-MR teaming. By integrating rule-based learning with experience-enhanced strategies such as in-context learning and retrieval-augmented generation, REBEL addresses several critical challenges, including user preference adaptation in MOO and situational awareness in dynamic team environments. Our extensive experiments validate REBEL's effectiveness across SOO, user preference alignment in MOO, and adaptation to situational changes. We also demonstrated REBEL's potential as a test-time adaptation module for pre-trained RL-based ITA policies. 

Future work will focus on extending REBEL's capabilities to handle more complex multi-agent environments, as long prompts describing large-scale missions pose comprehension challenges for the LLM. Additionally, integrating REBEL into real-world systems and refining its user preference modeling will provide deeper insights into its applicability across diverse and dynamic settings.


\typeout{}
\bibliography{main}

\begin{thebibliography}{10}
\providecommand{\url}[1]{#1}
\csname url@samestyle\endcsname
\providecommand{\newblock}{\relax}
\providecommand{\bibinfo}[2]{#2}
\providecommand{\BIBentrySTDinterwordspacing}{\spaceskip=0pt\relax}
\providecommand{\BIBentryALTinterwordstretchfactor}{4}
\providecommand{\BIBentryALTinterwordspacing}{\spaceskip=\fontdimen2\font plus
\BIBentryALTinterwordstretchfactor\fontdimen3\font minus \fontdimen4\font\relax}
\providecommand{\BIBforeignlanguage}[2]{{%
\expandafter\ifx\csname l@#1\endcsname\relax
\typeout{** WARNING: IEEEtran.bst: No hyphenation pattern has been}%
\typeout{** loaded for the language `#1'. Using the pattern for}%
\typeout{** the default language instead.}%
\else
\language=\csname l@#1\endcsname
\fi
#2}}
\providecommand{\BIBdecl}{\relax}
\BIBdecl

\bibitem{dahiya2023survey}
A.~Dahiya, A.~M. Aroyo, K.~Dautenhahn, and S.~L. Smith, ``A survey of multi-agent human--robot interaction systems,'' \emph{Robotics and Autonomous Systems}, vol. 161, p. 104335, 2023.

\bibitem{teaming2022state}
N.~A. of~Sciences~Engineering and Medicine, ``Human-ai teaming:state-of-the-art and research needs,'' \emph{National Academies of Sciences, Engineering and Medicine, Washington DC}, vol.~10, p. 26355, 2022.

\bibitem{yuan2025adaptive}
Z.~Yuan, R.~Wang, T.~Kim, D.~Zhao, I.~Obi, and B.-C. Min, ``Adaptive task allocation in multi-human multi-robot teams under team heterogeneity and dynamic information uncertainty,'' in \emph{2025 IEEE International Conference on Robotics and Automation (ICRA)}, 2025.

\bibitem{wang2023initial}
R.~Wang, D.~Zhao, and B.-C. Min, ``Initial task allocation for multi-human multi-robot teams with attention-based deep reinforcement learning,'' in \emph{2023 IEEE/RSJ International Conference on Intelligent Robots and Systems (IROS)}, 2023.

\bibitem{jo2023affective}
W.~Jo, R.~Wang, B.~Yang, D.~Foti, M.~Rastgaar, and B.-C. Min, ``Cognitive load-based affective workload allocation for multihuman multirobot teams,'' \emph{IEEE Transactions on Human-Machine Systems}, 2024.

\bibitem{wu2022task}
H.~Wu, A.~Ghadami, A.~E. Bayrak, J.~M. Smereka, and B.~I. Epureanu, ``Task allocation with load management in multi-agent teams,'' in \emph{2022 International Conference on Robotics and Automation (ICRA)}.\hskip 1em plus 0.5em minus 0.4em\relax IEEE, 2022, pp. 8823--8830.

\bibitem{wang2024husformer}
R.~Wang, W.~Jo, D.~Zhao, W.~Wang, A.~Gupte, B.~Yang, G.~Chen, and B.-C. Min, ``Husformer: A multimodal transformer for multimodal human state recognition,'' \emph{IEEE Transactions on Cognitive and Developmental Systems}, vol.~16, no.~4, pp. 1374--1390, 2024.

\bibitem{patel2020improving}
J.~Patel and C.~Pinciroli, ``Improving human performance using mixed granularity of control in multi-human multi-robot interaction,'' in \emph{2020 29th IEEE International Conference on Robot and Human Interactive Communication (RO-MAN)}.\hskip 1em plus 0.5em minus 0.4em\relax IEEE, 2020, pp. 1135--1142.

\bibitem{schmidbauer2023empirical}
C.~Schmidbauer, S.~Zafari, B.~Hader, and S.~Schlund, ``An empirical study on workers' preferences in human--robot task assignment in industrial assembly systems,'' \emph{IEEE Transactions on Human-Machine Systems}, vol.~53, no.~2, pp. 293--302, 2023.

\bibitem{fu2022robust}
B.~Fu, W.~Smith, D.~M. Rizzo, M.~Castanier, M.~Ghaffari, and K.~Barton, ``Robust task scheduling for heterogeneous robot teams under capability uncertainty,'' \emph{IEEE Transactions on Robotics}, vol.~39, no.~2, pp. 1087--1105, 2022.

\bibitem{lippi2023human}
M.~Lippi, J.~Gallou, J.~Palmieri, A.~Gasparri, and A.~Marino, ``Human-multi-robot task allocation in agricultural settings: a mixed integer linear programming approach,'' in \emph{2023 32nd IEEE International Conference on Robot and Human Interactive Communication (RO-MAN)}.\hskip 1em plus 0.5em minus 0.4em\relax IEEE, 2023, pp. 1056--1062.

\bibitem{chatzikonstantinou2020integrated}
I.~Chatzikonstantinou, I.~Kostavelis, D.~Giakoumis, and D.~Tzovaras, ``Integrated topological planning and scheduling for orchestrating large human-robot collaborative teams,'' in \emph{Conference on Biomimetic and Biohybrid Systems}.\hskip 1em plus 0.5em minus 0.4em\relax Springer, 2020, pp. 23--35.

\bibitem{zhang2016co}
C.~Zhang and J.~A. Shah, ``Co-optimizating multi-agent placement with task assignment and scheduling.'' in \emph{IJCAI}, 2016, pp. 3308--3314.

\bibitem{humann2018modeling}
J.~Humann and E.~Spero, ``Modeling and simulation of multi-uav, multi-operator surveillance systems,'' in \emph{2018 Annual IEEE International Systems Conference (SysCon)}.\hskip 1em plus 0.5em minus 0.4em\relax IEEE, 2018, pp. 1--8.

\bibitem{wang2024initial}
R.~Wang, D.~Zhao, A.~Gupte, and B.-C. Min, ``Initial task allocation in multi-human multi-robot teams: An attention-enhanced hierarchical reinforcement learning approach,'' \emph{IEEE Robotics and Automation Letters}, 2024.

\bibitem{humann2023modeling}
J.~Humann, T.~Fletcher, and J.~Gerdes, ``Modeling, simulation, and trade-off analysis for multirobot, multioperator surveillance,'' \emph{Systems Engineering}, vol.~26, no.~5, pp. 627--640, 2023.

\bibitem{yang2024driving}
R.~Yang, X.~Zhang, A.~Fernandez-Laaksonen, X.~Ding, and J.~Gong, ``Driving style alignment for llm-powered driver agent,'' in \emph{2024 IEEE/RSJ International Conference on Intelligent Robots and Systems (IROS)}.\hskip 1em plus 0.5em minus 0.4em\relax IEEE, 2024, pp. 11\,318--11\,324.

\bibitem{wang2024prefclm}
R.~Wang, D.~Zhao, Z.~Yuan, I.~Obi, and B.-C. Min, ``Prefclm: Enhancing preference-based reinforcement learning with crowdsourced large language models,'' \emph{IEEE Robotics and Automation Letters}, 2025.

\bibitem{mina2020adaptive}
T.~Mina, S.~S. Kannan, W.~Jo, and B.-C. Min, ``Adaptive workload allocation for multi-human multi-robot teams for independent and homogeneous tasks,'' \emph{IEEE Access}, vol.~8, pp. 152\,697--152\,712, 2020.

\bibitem{kannan2024smart}
S.~S. Kannan, V.~L. Venkatesh, and B.-C. Min, ``Smart-llm: Smart multi-agent robot task planning using large language models,'' in \emph{2024 IEEE/RSJ International Conference on Intelligent Robots and Systems (IROS)}.\hskip 1em plus 0.5em minus 0.4em\relax IEEE, 2024, pp. 12\,140--12\,147.

\bibitem{talebirad2023multi}
Y.~Talebirad and A.~Nadiri, ``{Multi-Agent Collaboration: Harnessing the Power of Intelligent LLM Agents},'' \emph{arXiv preprint arXiv:2306.03314}, 2023.

\bibitem{liang2023code}
J.~Liang, W.~Huang, F.~Xia, P.~Xu, K.~Hausman, B.~Ichter, P.~Florence, and A.~Zeng, ``Code as policies: Language model programs for embodied control,'' in \emph{2023 IEEE International Conference on Robotics and Automation (ICRA)}.\hskip 1em plus 0.5em minus 0.4em\relax IEEE, 2023, pp. 9493--9500.

\bibitem{lewis2020retrieval}
P.~Lewis, E.~Perez, A.~Piktus, F.~Petroni, V.~Karpukhin, N.~Goyal, H.~K{\"u}ttler, M.~Lewis, W.-t. Yih, T.~Rockt{\"a}schel \emph{et~al.}, ``Retrieval-augmented generation for knowledge-intensive nlp tasks,'' \emph{Advances in Neural Information Processing Systems}, vol.~33, pp. 9459--9474, 2020.

\bibitem{singh2023progprompt}
I.~Singh, V.~Blukis, A.~Mousavian, A.~Goyal, D.~Xu, J.~Tremblay, D.~Fox, J.~Thomason, and A.~Garg, ``{ProgPrompt: Generating Situated Robot Task Plans using Large Language Models},'' in \emph{2023 IEEE International Conference on Robotics and Automation}, 2023.

\bibitem{langchainEnsembleRetriever}
LangChain, ``Ensemble retriever,'' \url{https://python.langchain.com/v0.1/docs/modules/data_connection/retrievers/ensemble/}, 2024, accessed: 2024-09-16.

\bibitem{robertson2009probabilistic}
S.~Robertson and H.~Zaragoza, ``The probabilistic relevance framework: Bm25 and beyond,'' \emph{Found. Trends Inf. Retr.}, vol.~3, no.~4, p. 333–389, Apr. 2009.

\bibitem{cormack2009reciprocal}
Cormack \emph{et~al.}, ``Reciprocal rank fusion outperforms condorcet and individual rank learning methods,'' in \emph{Proceedings of the 32nd International ACM SIGIR Conference on Research and Development in Information Retrieval}, 2009, p. 758–759.

\bibitem{devlin2019bert}
J.~Devlin, M.-W. Chang, K.~Lee, and K.~Toutanova, ``Bert: Pre-training of deep bidirectional transformers for language understanding,'' 2019.

\bibitem{harriott2013modeling}
C.~E. Harriott and J.~A. Adams, ``Modeling human performance for human--robot systems,'' \emph{Reviews of Human Factors and Ergonomics}, vol.~9, no.~1, pp. 94--130, 2013.

\bibitem{watson2017informing}
M.~Watson, C.~Rusnock, M.~Miller, and J.~Colombi, ``Informing system design using human performance modeling,'' \emph{Systems Engineering}, vol.~20, no.~2, pp. 173--187, 2017.

\bibitem{pew1969speed}
R.~W. Pew, ``The speed-accuracy operating characteristic,'' \emph{Acta Psychologica}, vol.~30, pp. 16--26, 1969.

\end{thebibliography}
\bibliographystyle{IEEEtran}
\end{document}